\documentclass[a4paper]{article}
\usepackage{iwslt18,amssymb,amsmath,epsfig}

\usepackage{url}
\usepackage{todonotes}
\usepackage{booktabs}
\usepackage{multirow}

		\def\reg{{\rm\ooalign{\hfil
     \raise.07ex\hbox{\scriptsize R}\hfil\crcr\mathhexbox20D}}}

\newcommand\mn[1]{\textcolor{black}{#1}}
\newcommand{\mt}[1]{\textcolor{black}{ #1}}
\newcommand\ak[1]{\textcolor{black}{#1}}
\newcommand\sml[1]{\textcolor{black}{#1}}
\newcommand\short[1]{{}}

\title{Adapting Multilingual Neural Machine Translation to Unseen Languages} 
 \makeatletter
 \def\name#1{\gdef\@name{#1\\}}
 \makeatother
 \name{{\em Surafel M. Lakew$^{\dagger +}$ Alina Karakanta$^{\dagger +}$ Marcello Federico$^{+}$ Matteo Negri$^{+}$ Marco Turchi$^{+}$}}

\address{
$^{\dagger}$University of Trento, Trento, Italy \\ $^{+}$Fondazione Bruno Kessler, Trento, Italy \\ {\small \tt $^{\dagger}$name.surname@unitn.it, $^{+}$surname@fbk.eu}
}

\begin{document}
\maketitle

\begin{abstract}
Multilingual Neural Machine Translation (MNMT) for low-resource languages (LRL) can be enhanced by the presence of related high-resource languages (HRL), but the relatedness of HRL usually relies on 
\mn{predefined}
linguistic assumptions about language similarity.
\sml{
Recently, adapting MNMT to a LRL has shown to greatly improve performance. In this work, we explore the problem of adapting an MNMT model to an unseen LRL \ak{using data selection and model adaptation}.
}
\mn{In order to improve NMT for LRL,} we employ perplexity to select HRL data \mn{that are} most similar to the LRL on the basis of language distance.
We extensively explore data selection in popular multilingual NMT settings, namely in (zero-shot) translation, and in adaptation from a multilingual pre-trained model, for both directions (LRL$\leftrightarrow$en). 
We further show that dynamic adaptation of the model's vocabulary 
\mt{results in a more favourable segmentation for} the LRL in comparison with direct adaptation.
Experiments show reductions in training time and significant performance gains over LRL baselines, even with \textit{zero} LRL data (+13.0 BLEU), up to +$17.0$ BLEU for pre-trained multilingual model dynamic adaptation with related data selection. Our \mt{method}
outperforms current approaches, such as massively multilingual models and data augmentation, 
on four LRL.\footnote{
Scripts to replicate the experiments and pre-trained models: \\
\url{https://github.com/surafelml/adapt-mnmt}
}
\end{abstract}

\section{Introduction}
\label{Intro}
Neural Machine Translation (NMT) has become prevalent in the past years, contributing to the flow of information across languages and facilitating communication around the world. However, NMT requires a large amount of ``feature-label'' aligned data for building high-quality and usable
systems~\cite{koehn2017six}. For the majority of the world's languages, these resources are not available. Not benefiting from high quality MT (as it is usually the case with HRL) means that people's access to different sources of information can be restricted.

Multilingual Neural Machine Translation (MNMT) owes its success to cross-lingual knowledge transfer~\cite{Terence:LT}, which has been \mt{particularly} beneficial
for languages lacking large parallel data~\cite{johnson2016google}. Previous works 
\mn{document further improvements}
when using languages from the same family, however they all rely on 
\mn{predefined} linguistic assumptions about language similarity. 
Another challenge for facilitating access to information through MNMT is that relevant LRL data might not be available at the time of training the initial seed model, or not available at all. 
In most real-life applications, new needs in terms of domains or language coverage arise continuously, making monolithic MNMT models susceptible to out-of-vocabulary words. 
Moreover, new relevant training data in several (related or not) languages might become available continuously. Taking advantage of relevant data for adaptation is crucial to the performance of the final models~\cite{axelrod2011selection,wee2017selection}.

\sml{
Recently, building a large scale MNMT model was shown to be beneficial for LRL~\cite{aharoni2019massively}, \ak{even} outperforming \ak{models specifically fine-tuned on} the LRL data~\cite{neubig18:rapid}. Another approach optimizes embeddings through character n-grams (i.e., soft decoupled encoding, \ak{SDE)~\cite{wang2019SDE}}. A more recent data augmentation approach showed improvements over all the previous approaches by adapting the MNMT \ak{system} using pseudo-bitext generated by converting the HRL to the LRL~\cite{xia2019generalized}. Overall, research efforts in MT for LRL have shown that pre-training a multilingual NMT model and efficiently utilizing the available data are 
\mn{crucial towards} better translation quality.
}

\mn{In this paper, we investigate the usefulness of language similarity (distance between languages) as an indicator for selecting \textit{which} and \textit{how much} related HRL data can lead to the largest possible improvements.} 
\sml{
In 
\mn{analysing these aspects,}
we examine the potential of a pre-trained universal \mt{(MNMT)} model at two stages; {\it i}) without having access to the test language data at training time (zero-shot translation), and {\it ii}) after adapting it \ak{to the LRL} with selected data based on a language similarity criterion. We evaluate our hypothesis in the following proposed settings;
}

\textbf{Data Selection:} We 
\mn{compute the} perplexity of a LRL \ak{language} model on available HRL data, in order to choose HRL data 
\mn{that are most}
 similar to the LRL. Perplexity is a well-established information-theoretic measure, also used for measuring distance between languages \cite{gamallo2017pp}.
We evaluate the data selection technique in different scenarios; including a) language family, b) random, and c) our proposed perplexity-based selection criterion.

{\bf Training and Inference:}
First we examine the performance of the universal model in total absence of  LRL data (\textit{zero-shot}). 
The evaluation involves both translation directions (LRL$\leftrightarrow$en).
To date, model evaluation\mt{~\cite{neubig18:rapid,gu2018universal,wang2019SDE,xia2019generalized}} for the en$\rightarrow$LRL has not been  investigated yet. This direction is the most challenging \mn{one} because of the small amount of available target side data in the LRL and the morphological richness of several LRL compared to English. 

\textbf{Adaptation of Pre-Trained Model:} We experiment with 
\mn{the adaptation of} the multilingual NMT system by preserving the initial model vocabulary (\textit{DirAdapt}) or dynamically updating it to include new items (\textit{DynAdapt}), as in~\cite{lakew18:tl-dv}. Following previous observations that more frequent segmentation favors morphologically rich \ak{languages} and LRL~\cite{kreutzer2018char,cherry2018char}, we extend this approach by choosing different segmentation sizes that improve performance on the LRL.

Based on the above three aspects, this work aims at finding a viable way to improve a LRL translation task. The contributions of our work are three-fold\mn{. In particular, we}: 

\begin{itemize}
    \item \mn{Propose     an effective data selection method \ak{to select relevant data from several related HRL} that, on the same test languages, achieves better performance compared to the most recent data augmentation approach}.
    
    \item \ak{Explore the extreme case of a total absence of training data in the test language by attempting zero-shot translation \sml{using a model trained} with different portions of related HRL data in both translation directions.}
    
    \item Explore and compare approaches that aim to improve the quality of LRL translation, including direct and dynamic adaptation of \sml{ pre-trained models.}             \end{itemize}

\sml{
For a fair comparison with related works we utilize a standard 
\mn{dataset}
(TED Talks~\cite{qi2018andPre-trainedEmbd})
\mn{comprising} $58$ languages paired with English. 
Four languages are used as 
test. 
\mn{Two of them}
are 
extremely low-resource (Azerbaijani and Belarusian), while the other two (Galician and Slovak) are ``relatively'' low-resourced. 
We conduct our experiments using Transformer~\cite{vaswani2017attention}, which was shown to be superior for multilingual models~\cite{lakew2018comparison} and in HRL benchmarks~\cite{ott2018scaling}. Experimental results show the effectiveness of our 
\mn{approach, which outperforms those presented in previous works.}
}

\section{Adapting Multilingual NMT}
\sml{
In this work, we 
\mn{aim to find} a transfer-learning approach that leads to an efficient utilization of a pre-trained large-scale MNMT model. 
\mn{To achieve our goal of improving translation for the target LRL, we cast our approach as an unsupervised model adaptation strategy, in which relevant data for the adaptation are not supplied beforehand but have to be identified on the fly.}
}

\subsection{Data Selection by Language Distance}\label{approach:data-selection}
Perplexity is a commonly used measure to assess the quality of a language model~\cite{sennrich2012perplexity}, and has also been used to measure distance between languages \cite{gamallo2017pp}.\footnote{We propose perplexity over popular data selection techniques in domain adaptation \cite{axelrod2011selection,moorelewis2010}, because the large number of languages involved makes training pairwise language models unfeasible for the scope of this work.}
In this work, we use perplexity to select HRL data \mn{that are} 
similar to the LRL data. 
We train a language model on the LRL data ($LRL_{LM}$) and select training data with the lowest perplexity from related HRL ({\bf Select-pplx}). 
We compare this approach with: \begin{enumerate}
    \setcounter{enumi}{1}

    \item {\bf Select-one} -- Taking all available data only from one HRL related to the LRL. 

    \item {\bf Select-fam} -- Taking all available data from a set of HRL related to the LRL \mt{belonging to the same language family.}    
    \item {\bf Select-rand} -- \mn{Randomly sampling an equal proportion of data with Select-one and Select-pplx, from the HRLs that are closely related to the LRL.} \end{enumerate}
Perplexity is defined as the inverse probability of a test set (i.e., the HRL training data) computed using the $LRL_{LM}$. Thus, given the segments of the HRL set and the LM, the perplexity is computed as:

\begin{equation}
PP(S, LRL_{LM}) = \sqrt[\leftroot{-1}\uproot{1}N]{\prod_{i=1}^N \frac{1}{P(w_i | w_1^{i-1})}}
\label{eq1}
\end{equation}

Where: $S$ is 
\mn{a} HRL segment 
\mn{consisting the}
sequence $w_1, w_2, \ldots w_N$,  P($\cdot$) 
\mn{are}
the $n$-gram probabilities estimated on the training set of LRL$_{LM}$. The distance between the LRL and the HRL is computed by evaluating the $n$-gram of the latter using the $n$-gram model of the former. For each HRL set, consisting \mn{of} examples $S_j$, where $j = 1 \ldots m$,  we select $S_j$ with the lowest perplexity (i.e., closest to the LRL) by computing $PP(S_j, LRL_{LM})$. 
We repeat the process for each HRL, re-score the sentences of all HRL based on their perplexity and select the necessary portion of data determined by a pre-configured threshold.

\subsection{Direct vs. Dynamic Adaptation}\label{approach:dir-vs-dyn}  
For adaptation, we pre-process the test language data either {\it i}) using the pre-trained model's segmentation rules, or {\it ii}) by first learning a new segmentation model from the LRL data.
Thus, for the transfer-learning stage, we follow two strategies:
\begin{enumerate}
    \item {\bf DirAdapt}: Vocabularies, segmentation rules and all parameters of the pre-trained model are used without any change.
    
    \item {\bf DynAdapt}: New vocabularies are generated using the new segmentation rule, and portions of the pre-trained model parameter are re-used. 
\end{enumerate}

In the {\it DirAdapt} case, the segmentation rules of the pre-trained model are applied on the test language for 
the inference or adaptation stages.
In the {\it DynAdapt} case, rules are learned from the test language data and new vocabulary items are generated accordingly. \mt{At adaptation time, if the entries in the test language vocabulary are already present in the current dictionary, all the relative pre-trained model weights are transferred, while a random initialization of the embedding layers and the pre-softmax linear transformation weight matrix is performed for newly inserted vocabulary items.}
Unlike~\cite{lakew18:tl-dv}, we first look for the test language segmentation that maximizes the overlap with the pre-trained model vocabularies.\footnote{An alternative approach could be to remove the embedding and projection layers of the pre-trained model, however, preliminary results showed lower performance; thus avoided from the scope of this work.}

\begin{table}[!t]
    \centering
    \begin{tabular}{cc|ccc}     \\    gl 		& size 				& Lang 	& Select-fam 	& Select-pplx \\ \midrule
    Train 	& 10k 				& pt 	& 184k 	& 98.65k\\
    Dev  	& 682 				& es 	& 196k 	& 79.51k \\
    Test 	& 1,007 			& it 	& 204k 	& 6.85k \\\midrule
    & 		& \textbf{Total:} 	& 584k 	& 184k\\     \end{tabular}
    \caption{Data size for LRL gl, and selections using {\it Select-fam}, and {\it Select-pplx}.}
    \label{tab:size}
\end{table}

\subsection{Zero-shot Translation}\label{approach:zero-shot} 
We specifically aim at assessing the potential of the large scale MNMT model towards zero-shot \ak{translation (ZST)}. Unlike with adaptation strategies, the translation is evaluated in an extreme scenario, where the LRL has 
never been seen at training time. 
\mn{This means that} the transfer-learning to assist the LRL translation is expected to come from multiple languages, particularly related languages, that are present in the pre-trained model. We examine both a $LRL_{unseen}$$\leftrightarrow$$HRL$ translation directions, where:
\begin{enumerate}
    \item $LRL_{unseen}$ $\rightarrow$ HRL: represents a condition where the source side only sees related languages to the LRL, at \mt{training} time \ak{ but no LRL data at all}.
    
    \item HRL $\rightarrow$ $LRL_{unseen}$: 
        \mn{represents a so-far unexplored and more challenging condition, as discussed in Section \ref{Intro}.}
    
\end{enumerate}

To evaluate the two scenarios we pre-train several models with data 
\mn{featuring}
different size and language combinations. For constructing the data, we follow the perplexity-based data-selection criterion described in \S\ref{approach:data-selection}. In this proposal our objectives are; {\it i)} to evaluate how pre-trained models perform before an adaptation stage on \ak{unseen} test language data, and {\it ii)} how models trained 
\mn{on data with different levels}
of language relatedness behave in addressing a zero-shot translation.

Our expectation is that the more closely related language pairs (HRL) to the test language ($LRL_{unseen}$) are available, the 
\mn{higher the performance of the pre-trained models will be.}
Comparing the zero-shot translation against the adapted models using a similar data selection 
criterion and data of the $LRL_{unseen}$ will shade more light \mn{on} how much the pre-training helps. 
Moreover, the zero-shot translation can signal how robustly both the encoder and the decoder learn without seeing the test language, but different \ak{combinations} 
of related languages.

\section{Experiments}
\subsection{Data and Preprocessing}
\mn{For our experiments we use the TED talks corpus~\cite{qi2018andPre-trainedEmbd}, which contains  parallel data for $58$ languages aligned to English.}
\sml{As a first step, we use four LRLs paired with English (en) for evaluating the two adaptation strategies; 
including Azerbaijan (az), Belarusian (be), Galician (gl), and Slovak (sk)\ak{, and Turkish (tr), Russian (ru), Portuguese (pt), and Czech (cs) as their HRL respectively}.
All languages are used are used to train the massive MNMT model, except for the 
language serving as test language at each time.  
\ak{The choice of the test languages facilitates comparisons with previous works on similar settings.} As a second step, we select a single 
\mn{test language pair} \ak{(en$\leftrightarrow$gl)} for an in-depth analysis of the data selection strategies, the zero-shot inference and the adaptation approaches. The data set size of the four LRL with their linguistically closest HRL are used as in \cite{neubig18:rapid}.
}

Before each experiment, data is segmented into subword units using SentencePiece\footnote{https://github.com/google/sentencepiece}. We use the same pre-processing both for NMT and LM experiments. Following the recommendation \mn{in} \cite{denkowski2017stronger}, the segmentation rules are set to $8k$. 
\sml{The segmentation rules of the pre-trained models are used for the ZST and experiments using {\it DirAdapt}}. Unless otherwise specified, the same number of segmentation \mt{rules} is used for the {\it DynAdapt}, first by learning the rules using the test language.

\subsection{Measuring Language Distance}
\begin{figure}[!t]
    \centering
            \includegraphics[scale=.43]{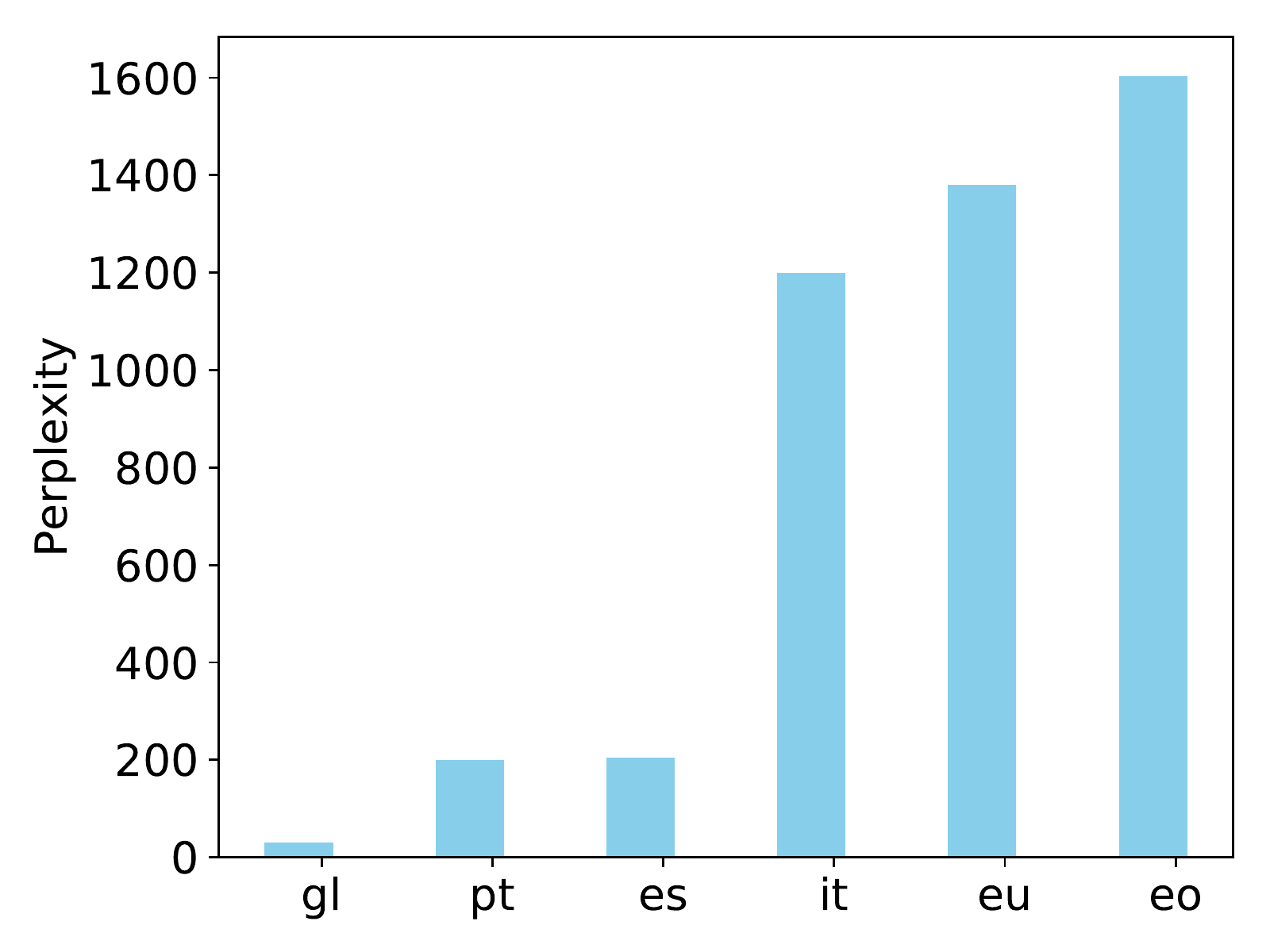}
    \caption{Perplexity for different languages (Portuguese/pt, Spanish/es, Italian/it, Basque/eu, Esperanto/eo) using the Galician/gl LM.}
    \label{fig:distance}
\end{figure}

\ak{For the related language data selection method,} we focus on one language, Galician (gl) as the test language, paired with Portuguese (pt), in addition to Spanish (es) and Italian (it) as further auxiliary languages. 
\sml{
First, we select pt as the closest language to gl ({\it Select-one}). \mn{Then, we include pt+es and pt+es+it for the experiments with selection based on the language family
({\it Select-fam}).}
}
For \textit{Select-pplx}, we train a neural language model\footnote{https://github.com/lverwimp/tf-lm} on the Galician data to re-score sentences from the training corpora of related languages and select the sentences with the lowest perplexity until we match the corpus size of Portuguese. Selection is made without replacement, i.e., an English sentence can have translations in multiple languages. \ak{ Statistics are shown in Table~\ref{tab:size}.}

As proposed in Section $\S$\ref{approach:data-selection}, \mn{the} distance between the test language to the rest of the languages is evaluated using \ak{perplexity of} the Galician LM against the test sets of all other languages. Figure~\ref{fig:distance} shows the closest languages.
The ranking reflects the proportion of each language in the mixed corpus with \mn{our} perplexity-based selection. Even though Galician is considered \mn{to be} more closely related to Portuguese, Spanish is behind \mn{it by} only $4$ perplexity points. Therefore, Spanish data is equally valuable for enhancing NMT performance on Galician. 

In order to check \mn{if} the improvement observed by adding more languages is due to simply having more training data, \sml{we set a maximum limit at time of data selection.} 
\mn{We hence} select the same amount of data (i.e., $184k$ \sml{for the case of Galician}) using \mt{each}
of the following approaches: Select-one, Select-rand, and Select-pplx.

\subsection{Model and Settings}
The LM \ak{used for data selection} is a $1$-layer LSTM model with embedding and hidden layer of $512$ units. We found that the best results were obtained when keeping all other settings as proposed in~\cite{verwimp2017tflm} (small model).
We train the translation models with the OpenNMT\footnote{http://opennmt.net/} TensorFlow implementation of the Transformer model~\cite{klein2017opennmt}. The model parameters are set to a $512$ hidden unit and embedding dimension, $4$ layers of self-attentional encoder-decoder with $8$ heads. At training time, we use $4096$ token level batch size with a maximum sentence length of $100$. For inference, we keep a $32$ example level batch size, with a beam search width of $5$. LazyAdam~\cite{kingma2014adam} is applied throughout all strategies with an initial learning rate constant of $2$. The learning rate increases linearly up to 
\mn{$8,000$ warm-up training steps,}
and decreases afterwards with an inverse square root of the training step. Given the sparsity of the test language data, dropout~\cite{srivastava2014dropout} is set to $0.3$. The pre-trained models are run for up to $1$M steps, 
\mn{and} the adaptations steps vary based on the amount of data used. 
In all runs, models are observed to converge.

\begin{table*}[!t]
\centering
\begin{tabular}{clcccc|r}
 & Strategy & az[tr] & be[ru] & gl[pt] & sk[cs] & AVG. \\ \midrule
\multirow{3}{*}{Neubig \& Hu 2018} & Baseline & 2.70 & 2.80 & 16.20 & 24.00  & 11.43 \\ \cmidrule{2-7}  & MNMT$\rightarrow$Bi (RapAdapt) & 10.70  & 17.40 & 28.40 & 28.00 & 21.20 \\ Wang et al., 2018 & MNMT$\rightarrow$Bi (SDE) & 11.82  & 18.71 & 30.30 & 28.77  & 22.40 \\ \cmidrule{3-7}
 & $\Delta$ (SDE-Baseline) & 9.12  & 15.91  & 14.10 & 4.77 & 10.98 \\ \midrule
Aharoni et al., 2019 & Many $\leftrightarrow$ Many & 12.78 & 21.73 & 30.65 & 29.54  & 23.67\\ \midrule
Xia et al., 2019 & Data-Augment & {\bf 15.74}    & {\bf 24.51} & 33.16 & 32.07 & 26.37\\ \midrule
\multirow{4}{*}{Ours} & Baseline & 3.61  & 4.42  & 16.32  & 26.44 & 12.70 \\ \cmidrule{2-7}
 & MNMT$\rightarrow$Bi (DirAdapt) & 14.43 & 22.06  & 33.53 & 30.13 & 25.04 \\  & MNMT$\rightarrow$Bi (DynAdapt) & 15.33 & 23.80 & {\bf 34.18} & {\bf 32.48} & {\bf 26.45} \\ \cmidrule{3-7}
& $\Delta$ (DynAdapt-Baseline) & {\bf 11.72}  & {\bf 19.38}  & {\bf 17.86} & {\bf 6.04} & {\bf 13.75} \\ 
\bottomrule
\end{tabular}
\caption{BLEU scores for the four LRL$\rightarrow$en comparing against previous approaches; RapAdapt~\cite{neubig18:rapid}, SDE~\cite{wang2019SDE}, Many$\leftrightarrow$Many~\cite{aharoni2019massively}, and Data-Augment~\cite{xia2019generalized}. Bi is an adaptation with the LRL + [closest-HRL] according to {\it Select-one} strategy.}
\label{tab:comparison}
\end{table*}

\subsection{Baselines and Comparison}
Single language pair models \ak{(baselines)} are trained from scratch using only the test LRL data. First, results from the adaptation and data-selection strategies are compared with these baselines. 
\mn{Then,}
we compare against \mn{solutions previously proposed in literature, namely:} 
\begin{itemize}
    \item A direct adaptation of \mn{the} multilingual model to the LRL, {\it RapAdapt}~\cite{neubig18:rapid} and {\it SDE}~\cite{wang2019SDE}.     
    \item A massive multilingual model trained including all the test LRL, avoiding adaptation ({\it Many} $\leftrightarrow$ {\it Many})~\cite{aharoni2019massively}.
    
    \item A data-augmentation for LRL pair, followed by adaptation of a multilingual model ({\it Data-Augment})~\cite{xia2019generalized}.
\end{itemize}

In the first \mn{two} cases ({\it RapAdapt, SDE}), a similar strategy to our {\it DirAdapt} is implemented using an RNN model. For a fair comparison with our Transformer-based approach, we take the \ak{relative improvement (}$\Delta$) between the single pair baselines and the dynamically adapted models. The second (Many$\leftrightarrow$Many) and third ({\it Data-Augment}) approaches utilize the Transformer model, allowing us to directly compare against \sml{the reported results.} More interestingly, these comparisons bring together several approaches using the same four test languages, aiming at improving the quality of LRL translation. As a metric to evaluate translation quality, we use BLEU~\cite{papineni2002bleu}.

\section{Results and Analysis}
\subsection{Adaptation Does Matter}
In Table~\ref{tab:comparison}, we show the $\Delta$ between the baseline \ak{of~\cite{neubig18:rapid}} ({\it RapAdapt}) and \mn{the} best performing adaptation approach (\textit{SDE}), against the $\Delta$ between our baseline and our best performing approach (\textit{DynAdapt}). Even with stronger baselines, our $\Delta$ is higher than in the previous approaches with $+2.77$ BLEU averaged over the four test languages. Note that the MNMT model refers to a training setting with all except for the test language \ak{(cold start)}. We also note that our {\it DirAdapt} outperformed the {\it RapAdapt} and {\it SDE} with a larger margin in all \mn{the} test languages.

The authors in~\cite{aharoni2019massively} argue that the better performance of {\it Many$\leftrightarrow$Many} over the {\it RapAdapt} and {\it SDE} 
\mn{is due} to avoiding model over-fitting by including more languages on both the encoder and decoder sides. However, our adaptation strategies show better performance in all test cases, with a $+1.37$ ({\it DirAdapt}) and $+2.78$ ({\it DynAdapt}) average BLEU. In fact, the additional improvement from {\it DirAdapt} comes from curating the segmentation for the test language and partially transferring the MNMT model parameters.

By contrasting the performance of previous works against the {\it DynAdapt}, we learn that our method is superior 
\mn{to all, in average BLEU score}. Specifically, when compared to the latest {\it Data-Augment}~\cite{xia2019generalized}, the {\it DynAdapt} shows better performance in two of the test languages (gl, sk), and slight degradation for az and be. Our observation for the lower performance is that the data augmentation results in much larger synthetic data, while our adaptation utilized only the original LRL data for each of the test languages and the closest related language pair (amounting to a max of $~200k$ segments) as in~\cite{neubig18:rapid}. Overall, our approach showed the possibility of achieving better performance when initializing from pre-trained MNMT parameters.

\subsection{Zero-shot Translation}
Comparing the approaches in~\cite{neubig18:rapid} that used RNNs for evaluating the ZST settings against our results, we observe a large difference (see Table~\ref{tab:results_zst_all}) that again attests the superiority of the Transformer model. The better performance is particularly true for the MNMT models that are trained using all the available data but the test language. Previous works have also shown similar findings for the Transformer model when it comes to zero-shot translation~\cite{lakew2018comparison,aharoni2019massively}. Thus, it is important to emphasize \mn{that} the multilingual model is the best suit for further investigation by applying the data-selection procedures with the two adaptation options.

\begin{table}[!t]
\small
\centering
\begin{tabular}{clcccc}
 & Strategy & az & be & gl & sk \\ \midrule
\multirow{2}{*}{Neubig \& Hu} & Select-one    & 3.80 & 2.50 & 8.60 & 5.40  \\
                                                    & MNMT  & 3.70 & 3.50 & 15.50 & 7.30 \\ \midrule
\multirow{2}{*}{Ours}                               & Select-one    & 3.25 & 2.07 & 13.59 & 9.30 \\ 
                                                    & MNMT  & {\bf 11.06} & {\bf 10.97}   & {\bf 27.28}  & {\bf 20.57} \\
\bottomrule
\end{tabular}
\caption{Results for LRL$\rightarrow$en ZST using model trained with a single pair {\it Select-one}, and all but the test LRL ({\it MNMT}).}
\label{tab:results_zst_all}
\end{table}

\subsection{Data Selection for Zero-shot translation}
Table~\ref{tab:results_zst} shows results for ZST using various data-selection strategies.
In the gl$\rightarrow$en direction, adding more data from related languages improves performance but the improvement slows down as more languages are added. Even without any test language data, performance increases from $13.59$ BLEU for training only with pt ({\it Select-one}) to $24$ \mn{BLEU} for pt+es ({\it Select-fam}), while with it \ak{(pt+es+it)} increases further by $1.34$ BLEU. The MNMT model scores higher, but only by $1.77$ BLEU when compared to best {\it Select-fam} strategy. Here, it is important to \mn{emphasize that:}
{\it i}) the MNMT model is trained using over $5$M segments except for the test (gl-en) pair, meaning \mn{that} the performance of {\it Select-fam} shows the possibility to improve a ZST by having more related languages but less data, {\it ii}) while the amount of data is the same for {\it Select-one}, {\it Select-rand}, and {\it Select-pplx}, the latter shows better performance, \mn{indicating} the importance of the data selection criteria using perplexity.  

Opposite results are obtained in the en$\rightarrow$gl direction. As expected, our second evaluation of ZST into an unseen language on the decoder side does not perform well \ak{and performance decreases as more languages are added (i.e. from pt+es to pt+es+it)}. However, selecting related-language data using \textit{Select-pplx}, we observe a better performance among the data-selection criteria at $5.38$ BLEU. 

Overall, ZST performance when translating from an unseen source language (gl) into a seen target language (en) is better than the baseline (see Table~\ref{tab:results_zst}), with more than $10.0$ BLEU points.  This gain highlights the importance of closely related languages for improving the performance on the LRL. However, the opposite \ak{direction,} where we infer into unseen target language (gl), is a more challenging task that 
\mn{requires} further investigation and the availability of at least monolingual data for the LRL.

\begin{table}[!t]
\small
  \centering
  \begin{tabular}{cccc}     & Strategy & gl$\rightarrow$en & en$\rightarrow$gl \\ \midrule
   Our non ZST 				& Baseline 		& 16.32 & {\bf 11.83} \\ \midrule
  							& Select-one       & 13.59 & 8.05 \\   \multirow{4}{*}{Ours ZST} & Select-rand 	& 14.69 & 4.09 \\
  						 	& Select-pplx 	& 15.55 & 5.38 \\ \cmidrule{2-4}
  						 	& pt$+$es 		& 24.17 & 4.61 \\
  						 	& pt+es+it 		& 25.51 & 4.17 \\ \cmidrule{2-4}
  							& MNMT          & {\bf 27.28} & {\bf 8.78} \\ 
  \bottomrule
  \end{tabular}
  \caption{BLEU for ZST using models trained with different data-selection criteria. Pt+es and pt+es+it are the two varieties of the {\it Select-fam} method.   }
  \label{tab:results_zst}
\end{table}

\subsection{Data Selection for Adaptation}
\begin{table}[!t]
\small
  \centering
  \begin{tabular}{lcc}   MNMT Adaptation & gl$\rightarrow$en & en$\rightarrow$gl \\ \midrule
   Strategy             & Dir/Dyn- Adapt   & Dir/Dyn- Adapt \\ \midrule
   $\rightarrow$ {\it gl} 				& 32.18 / -2.5 		& 26.39 / -3.21 \\ \cmidrule{2-3}
  $\rightarrow$Select-one + {\it gl} 			& 33.53 / +0.65 	& 26.45 / +0.28 \\
  $\rightarrow$Select-rand + {\it gl} 	& 32.61 / +0.75 	& 25.94 / +0.06\\ \cmidrule{2-3}
  $\rightarrow$Select-pplx + {\it gl} 	& $^{\uparrow}${\bf 34.15} / $^{\uparrow}${\bf +1.41} & $^{\uparrow}${\bf 27.35} / $^{\uparrow}${\bf +0.59}\\ \cmidrule{2-3}
  $\rightarrow$pt+es+it + {\it gl}		& 33.38 / +2.14 	& 26.40 / +1.14\\
  \bottomrule
  \end{tabular}
  \caption{BLEU using models adapted from the MNMT in different data conditions. $^{\uparrow}$ indicates statistically significance using bootstrap re-sampling ($p<0.05$)~\cite{Koehn2004}.
  }
  \label{tab:results}
\end{table}

\ak{Table~\ref{tab:results} shows results for adapting a MNMT model with data selected using our proposed perplexity-based method, both in the direct and the dynamic adaptation scenario. As a general rule, adaptation with selecting data from several languages improves over adapting only with the target language. One possible reason for this improvement is avoiding overfitting to the little data of the target language, as shown in~\cite{neubig18:rapid}. However, perplexity-based data selection (\textit{Select-pplx}) outperforms selecting only one related language (\textit{Select-one}) for both translation directions. Moreover, we show that the improvement does not come only from mixing several related languages, since \textit{Select-rand} hurts performance for both directions. Our method improves even over adapting with all data from the most related languages (\textit{pt+es+it}), allowing for a faster adaptation.}

\begin{figure}[!t]
    \centering
            \includegraphics[scale=.47]{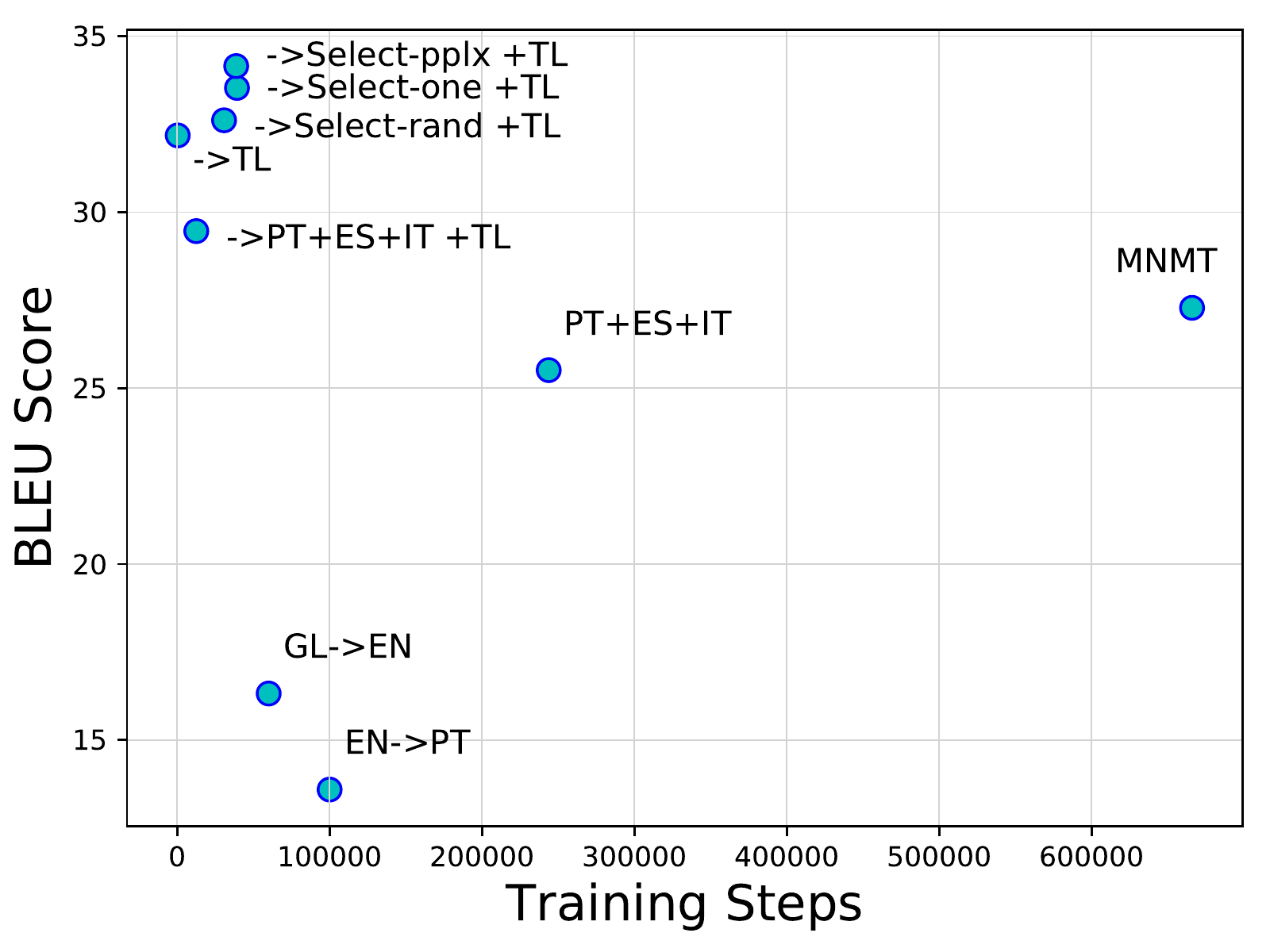}
        \caption{BLEU vs. training steps for gl$\rightarrow$en (TL) direction.}
    \label{fig:train-step_vs_bleu}
\end{figure}

The results show that perplexity can be a reliable measure for selecting smaller amounts of related-language data both in \ak{translation and adaptation from a MNMT model} in order to obtain larger improvements, with reduced training time (see Figure~\ref{fig:train-step_vs_bleu}).
For data selection strategies (either with perplexity or random), better performance is achieved with faster convergence. This confirms that the data-selection and the adaptation strategy is the fastest way to build a usable and better performing system for an unseen language from a pre-trained model.

Comparing the results of {\it DirAdapt} and {\it DynAdapt}, {\it DynAdapt} shows consistent improvements \ak{when adaptation is performed with data from at least two languages (LRL+another language)}. 
\ak{This can be attributed to the fact that} the {\it DirAdapt} has a complete overlap ($100$\%) both for the source and target side vocabularies with the pre-trained initial model (i.e., the initial model vocabulary is used without any modification), as well as the transfer of all parameters when adapting. \ak{On the contrary}, the {\it DynAdapt} improvement comes from a careful segmentation of the test language before adaptation, resulting in a new vocabulary and consequently enforcing a partial transfer of parameters from the initial model. In addition to the importance of data-selection, the additional gain using {\it DynAdapt} 
\mn{indicates that} a universal multilingual model can be 
\mn{made stronger}
if tailored to the characteristics of the test languages when adapting.

\mn{When} conducting a qualitative evaluation of the segmentation, for extremely low-resource test languages (such as gl$\leftrightarrow$en with $10k$ and az$\leftrightarrow$en with $5k$ bitext), we observed a frequent segmentation that favours sub-words closer to character level for most of rare words included in the vocabulary. This is consistent 
\mn{with}
previous work supporting character level segmentation for improving NMT of LRL~\cite{kreutzer2018char,cherry2018char}. Furthermore, with a reduced vocabulary size, {\it DynAdapt} can compress the model with smaller embedding and pre-softmax linear transformation dimensions compared to the pre-trained model, and \ak{with} sharing all the updated weight matrix as in~\cite{press2016using}.

\section{Related Work} 
Multilingual NMT approaches share a common feature by aggregating data from various language pairs. In comparison with earlier approaches~\cite{dong2015multi,luong2015multi,firat2016multi}, training a single attentional encoder-decoder (universal) model using multiple pairs showed to be an efficient multilingual setting~\cite{johnson2016google, ha2016toward}. While the performance of a universal model for HRL is comparable with a strong single language pair baseline, LRL pairs gain the highest improvement from the cross-lingual transfer. 
Thus, transfer learning for LRL can be defined in two main forms; {\it i}) ``vertically'', aggregating data from several language pairs to train a single model~\cite{johnson2016google}, {\it ii}) ``horizontally'', pre-training a model with the available pairs and fine-tuning it using the test (LRL) language data~\cite{zoph16:tf,nguyen2017transfer,lakew18:tl-dv}, or {\it iii}) with a combination of the two approaches.

Recently, new approaches have been introduced to efficiently adapt a pre-trained model to a LRL. One such case is proposed by~\cite{neubig18:rapid}, where they suggest to train a universal model (i.e., a model trained using up to 58 LRL-en pairs), with or without the test language direction. At time of adaptation, first, they adapt using only the LRL-en pair. Alternatively, a closely related language pair is added to the LRL-en as a regularizer when adapting. Both of the adaptation strategies show a larger performance gain over baseline models trained from scratch, however, their findings show the latter as an optimal adaptation setting.

Aimed at improving the source side language representation and parameter sharing,~\cite{wang2019SDE} introduced a multilingual lexicon encoding through character embedding, called Soft Decoupled Encoding. Their approach shows better performance than the adaptation strategies in~\cite{neubig18:rapid}, using a similar evaluation pairs. In a different work, a many-to-many multilingual model training is explored using all the available pairs both in LRL$\leftrightarrow$en directions~\cite{aharoni2019massively}. By avoiding the adaptation stage, the approach showed to perform better when compared to the results in~\cite{neubig18:rapid,wang2019SDE} that utilize a many-to-one setting. 

More recently, a data augmentation strategy is proposed to further improve the LRL pairs~\cite{xia2019generalized}. The approach leverages a target side monolingual and closely related HRL-English parallel data. Back translation is used to generate a pseudo-HRL from the monolingual data, while the HRL side of the parallel data is converted to pseudo-LRL using word substitution from a bilingual dictionary, similar to the approach in~\cite{Karakanta2018}. 
The synthetic data is used to construct a pseudo-HRL-en and a pseudo-LRL-en pair. Then, the synthetic data together with the available small LRL-en test language is used to improve over the baseline models.
Using the same test languages, the data augmentation approach, which creates additional parallel data for the adaptation stage, outperformed the approaches reported in previous works~\cite{neubig18:rapid,wang2019SDE,aharoni2019massively}.

This work shares a common ground on the effectiveness of pre-training a universal model and adaptating it to ultimately improve LRL pairs, however, it differs on the following aspects: {\it i}) it only considers a scenario where all of the pre-trained models have never seen the test language pair, {\it ii}) it learns a language model on the LRL to select data from related languages, {\it iii}) it investigates the less explored direction of en-LRL translation, {\it iv}) it explores zero-shot translation without adapting the pre-trained model, and {\it v}) it extends the dynamic adaptation strategy~\cite{lakew18:tl-dv}, that customizes any pre-trained model to the LRL pair.

In general, aggregating related HRL pair data with the LRL for an adaptation stage showed to perform better in all the test cases. Unlike in~\cite{neubig18:rapid}, who utilize only the immediately related language, we chose segments from different related languages based on the perplexity measure. Moreover, our approach does not rely on additional monolingual data or augmentation as in~\cite{xia2019generalized}, instead, efficiently utilizes multiple related languages by identifying the relevant examples to the test language pair.

\section{Conclusion}
In this work, we focused on enhancing NMT performance for LRLs with data selection, and direct and dynamic adaptation of pre-trained models. To this aim, we 
\mn{used}
perplexity to select the most relevant data to the test language. We show that perplexity-based data selection improves translation, \sml{leading to an improvement} 
 of up to 10.0 BLEU points for LRL$\rightarrow$en and 17.0 BLEU points for en$\rightarrow$LRL when adapting from a multilingual model, \ak{with reduced training time}. Our adaptation strategy with selected data is useful even in the extreme case of zero-shot translation for \ak{an} unseen language (+13.0 BLEU). 
In future works, we plan to integrate our approach with data augmentation and semi-supervised model training strategies. 

\bibliography{iwslt2019}
\bibliographystyle{IEEEtran}

\end{document}